\documentclass[runningheads]{llncs}
\usepackage[T1]{fontenc}
\usepackage{graphicx}
\usepackage[dvipsnames]{xcolor}

\usepackage{amsmath}
\usepackage{mathtools}
\usepackage{epic}
\usepackage{color}

\definecolor{azul}{rgb}{0.1,0.2,0.7}
\definecolor{verde}{rgb}{0.4,0.6,0.4}
\definecolor{bordo}{rgb}{0.8,0.3,0.3}
\definecolor{rojo}{rgb}{0.8,0.3,0.3}
\definecolor{gris}{rgb}{0.4,0.4,0.4}
\definecolor{amarillon}{RGB}{245,226,190}
\definecolor{amarillon1}{RGB}{253,224,180}
\definecolor{amarillon2}{RGB}{247,250,210}
\definecolor{naranja}{RGB}{255,150,102} 
\usepackage{graphicx}               

\usepackage{verbatim}
\definecolor{azul}{rgb}{0.1,0.2,0.6} 
\definecolor{verde}{rgb}{0.1,0.5,0.3}
\definecolor{bordo}{rgb}{0.7,0.3,0.3}
\usepackage{lipsum}
\usepackage{slashed}
\usepackage{ragged2e}
\usepackage[T1]{fontenc}
\usepackage[utf8]{inputenc}
\usepackage{eufrak}
\usepackage{isomath}
\usepackage{upgreek}
\usepackage[export]{adjustbox}
\usepackage{mathrsfs}   
\usepackage[scr=dutchcal,calscaled=1]{mathalfa}
\usepackage{relsize} 
\usepackage{amsmath,amssymb}
\usepackage{stmaryrd} 

\usepackage{tikz}
\usetikzlibrary{automata}
\usetikzlibrary{topaths}
\usepackage{pstricks}
\usetikzlibrary{automata,positioning,calc}
\usetikzlibrary{arrows}
\usepackage{xcolor}
\usepackage[basic]{circ}
\usepackage{graphics}
\usepackage{cancel}

\usepackage{soul}

\usepackage{relsize} 
\usepackage{amsmath,amssymb}
\usepackage{tikz-cd}
\usetikzlibrary{automata}
\usetikzlibrary{topaths}
\usepackage{pstricks}
\usetikzlibrary{automata,positioning,calc}
\usetikzlibrary{arrows}
\usepackage{circuitikz}
\usepackage{tikz}
\usetikzlibrary{mindmap,trees}
\usepackage{color,xcolor}

\begin{document}
\title{Recurrent Neural Networks as Electrical Networks, a formalization}
%
%
\author{Mariano Caruso\inst{1,2}\orcidID{0000-0002-7455-1193}\\ \and
Cecilia Jarne\inst{3,4}\orcidID{0000-0002-3917-920X} }
\authorrunning{M. Caruso and C. Jarne}
%
\institute{Fundación I+D del Software Libre, FIDESOL \and Facultad de Ciencias, Universidad de Granada, España \email{mcaruso@fidesol.org} \\ \and 
Universidad Nacional de Quilmes UNQ - Departamento de Ciencia y Tecnolog\'ia \and CONICET \email{cecilia.jarne@unq.edu.ar}
}
\maketitle              
%


\begin{abstract}

 Since the 1980s, and particularly with the Hopfield model, recurrent neural networks or RNN became a topic of great interest. The first works of neural networks consisted of simple systems of a few neurons that were commonly simulated through analogue electronic circuits. The passage from the equations to the circuits was carried out directly without justification and subsequent formalisation. The present work shows a way to formally obtain the equivalence between an analogue circuit and a neural network and formalizes the connection between both systems. We also show which are the properties that these electrical networks must satisfy. We can have confidence that the representation in terms of circuits is mathematically equivalent to the equations that represent the network.

\end{abstract}

\section{Introduction}

During the 1980s, and particularly since the Hopfield model \cite{Hopfield3088}, recurrent neural networks became a topic of great interest. In particular, latter with the works of Funahashi, Nakamura and Kurita \cite{DBLP:journals/nn/Funahashi89,DBLP:journals/nn/FunahashiN93,DBLP:journals/nn/KuritaF96}, which made it possible to link neural networks with the description of dynamic systems, this research field began to establish as an area in itself. There are multiple papers on how neural networks are universal approximators, (i.e. they can approximate any continuous function). The proofs tell that neural networks can approximate any continuous function \cite{10.1007/11840817_66,10.1145/130385.130432}. Any finite-time trajectory of a given $n-$dimensional dynamical system can be approximately realized by the internal state of the output units of a continuous-time recurrent neural network with $n-$output units, some hidden units, and an appropriate initial condition \cite{DBLP:journals/nn/FunahashiN93}. It was not until the last ten years that the current computing algorithms, the new hardware and the theory of neural networks allowed enormous developments in various areas related to natural language, dynamical systems, neurosciences and time series analysis using such networks \cite{TRISCHLER201667,Gerstner60}.


However, with the hardware that was available at that time, the first works of neural networks consisted of simple systems of a few neurons that were commonly simulated through analogue electronic circuits. The passage from the equations to those circuits in most of the works of  that time was carried out in a direct way, without much formalism. Mainly because the objective was to show how effectively these circuits could constitute rudimentary neural networks, mind that the analogue circuits allowed to simulate these systems. The behaviour of the systems was studied when the parameters of the network varied. Some current works in the area of electronics take up this idea and through circuit simulations, or the synthesis of analogue circuits, study the properties in systems with few neurons.  One example are the transitions to chaotic systems \cite{Ansari2011,TabekouengNjitacke2020}.

On the other hand, dedicated circuits are used in another field called Neuromorphic engineering \cite{58356,10.1145/2601069}, which is also known as neuromorphic computing. This is the use of very large-scale integration (VLSI) systems containing electronic analogue circuits that mimic neuro-biological architectures related to the nervous system. A neuromorphic computer is called any device that uses physical artificial neurons to do computations. Recently the term neuromorphic has been used to describe analogue, digital, mixed-mode analogue/digital VLSI (and also software systems) that implement models of neural systems used to understand perception, motor control, or multisensory integration. The implementation of neuromorphic computing on the hardware level is realized through oxide-based memristors, spintronic memories, threshold switches, and transistors. Training software-based neuromorphic systems of spiking neural networks can be achieved using error backpropagation, e.g., using Python-based frameworks.  The training algorithms of these systems are still complex, and it is difficult to control the network parameters.

Motivated by these developments and the gap in the literature about the formal aspects, the present work shows how to formally obtain the equivalence between an analogue circuit and a neural network, and formalizes the connection between both systems. We also show which are the properties that these electrical networks must satisfy. To the best of our knowledge, this is not explicitly found in the literature. The aim of the analysis is to explain the case of the linear network, meaning when the transfer function is the identity. This case is of interest since it has been used in various works, but also because it is often used to approximate nonlinear systems to first-order \cite{DBLP:journals/neco/SussilloB13,SUSSILLO2014156}.

\section{Notation and dynamics}

We have a set of $n-$\textit{artificial neurons}, for each of these there is a dynamic quantity called \textit{activity}, represented by a function $h_i:\mathscr{T}\longrightarrow \mathscr{H}\subseteq\mathbb{R}$ with $i{\in}I_n$ and a  $\mathscr{T}\subseteq \mathbb{R}$ is the set of temporary labels. We will use the following compact notation to denote the discrete set $n$ natural numbers $I_n{=}\{1,\cdots, n\}{\subset} \mathbb{N}$. We can arrange these $n$ functions in a column vector $\pmb{h}{=} (h_1,\cdots,h_i,\cdots, h_n)^{\mathfrak{t}}$, where $\mathfrak{t}$ denotes matrix transposition. The vector $\pmb{h}$ represents the \textit{state} of the activity of the network (formed by the $n$ neurons) at that time $t$. On the other hand, there are a series of $m$ input functions, $x_k:\mathscr{T}\longrightarrow \mathscr{X}\subseteq \mathbb{R}$ with $k{\in}I_m$, which can be arranged in a column vector $\pmb{x} = (x_1,\cdots,x_k,\cdots, x_m)^{\mathfrak{t}}$. For recurrent neural networks (RNN) the activity vector $\pmb{h}$ satisfies:
\begin{equation}\label{din red}
\pmb{\dot{h}}(t)=-\pmb{\lambda}\,\pmb{h}(t) + \pmb{\sigma}\pmb{(}\pmb{w\,h}(t)+\widetilde{\pmb{w}}\,\pmb{x}(t)\pmb{)},
\end{equation} 
where $\pmb{\dot{h}}$ represents the total derivative with respect to time in the usual sense. The diagonal matrix $\pmb{\lambda}$ contains the inverses of the characteristic times ($\tau_k$, $k{\in} I_n$), of the postsynaptic modes of each neuron, $\pmb{\lambda}^{-1}=diag(\tau_1,\cdots,\tau_k,\cdots,\tau_n)$. In the case where the neural network is completely disconnected, both internally $\pmb{w}=\pmb{0}$ and externally $\widetilde{\pmb{w}}=\pmb{0}$, the activity of each neuron $h_k$ of the network decays exponentially each with a characteristic time given by $\tau_k$. The matrices $\pmb{w}$ and $\widetilde{\pmb{w}}$ are $n\times n$ and $n\times m$, respectively. The matrix elements $\pmb{w}$, $w_{ij}$, contain the synaptic connections, similar to $\widetilde{\pmb{w}}$, $\pmb{\sigma}:\mathbb {R}^n \longrightarrow \mathbb{R}^n$ is a vector field of \textit{activation}. Strictly speaking, these fields usually have their image in some compact set, since the activation of the neurons has a saturation behavior, the typical examples are \textit{hyperbolic tangent}, \textit{logistics}, etc., which satisfies $\pmb{\sigma}(\pmb{0})=\pmb{0}$ (because the activity cannot be revived instantly, that is, the result of activating a neuron with zero activity is null). Furthermore, each of its components is defined by applying a single nonlinear function $\sigma:\mathbb{R}\longrightarrow\mathbb{R}$. That is, given a vector $\pmb{\xi}\in \mathbb{R}^n$, expressed in components as $\pmb{\xi}{=}(\xi_1,\cdots,\xi_n)$ is has to $\pmb{\sigma}(\pmb{\xi})=\pmb{(}\sigma(\xi_1),\cdots,\sigma(\xi_n)\pmb{)}$, as usual. The activity state of the network is determined by \eqref{din red}, which is updated as a result of the interaction between them via $\pmb{w}$, with the external signals $\pmb{x}$ that intervene on the activity of neurons according to $\widetilde{\pmb{w}}$, and together with some initial condition. We could write \eqref{din red} compactly as:
\begin{equation}\label{din redd}
\pmb{\dot{h}}(t)=\pmb{F}\pmb{(}\pmb{h}(t),\pmb{x}(t)\pmb{)}.
\end{equation} 
There are two procedures that can be performed on this differential equation in order to say something about the behavior of the system  models: discretization and linearization. The first procedure allows computing the model through an algorithm. The second will allow us to clearly find a linear electrical network that captures all the dynamics of the recurrent neural network. The order of these procedures does not alter the result, or in other words, the order in which they are applied is independent. Intuitively we can anticipate this result, since each procedure is introduced by a different member of \eqref{din redd}, discretization is applied on the differential operator on the left, while linearization is done on the activation function on the right.

\section{Linearization process}


As the nonlinear character of $\pmb{F}$ is exclusively in $\pmb{\sigma}$, linearizing is a procedure that has to do with the \textit{activation field} and the neuronal activity itself, that is, the activation object, which we are interested in considering. Within this activation field $\pmb{\sigma}$, let us now look at each function within $\sigma:\mathbb{R}\longrightarrow \mathbb{R}$, and suppose that $\sigma(\xi)$ is $k-$times differentiable at $\xi=0$, by Taylor's theorem, there is a function \textit{remainder} $R_k(\xi)$ that allows us to write as 
\begin{equation}
\sigma(\xi)=\sigma(0)+\pmb{(}d_\xi\sigma(\xi)|_{\xi=0}\pmb{)} \xi +\cdots+\pmb{(}d_\xi^{(k)}\sigma(\xi)|_{\xi=0}\pmb{)} \xi^k+ R_k(\xi)\xi^{k}.
\end{equation} 
Activation functions $\sigma$ are usually chosen whose tangent line has a slope equal to $1$ at $\xi=0$, that is, the activation function near the origin resembles the identity function, let us remember that $\sigma(0) =0$. Using all this and for $\xi$ small enough we can approximate $\sigma(\xi)\simeq \xi$. 
Local linear approximations also represent an important building
block for the analysis of the behaviour of more complex,
nonlinear dynamical systems \cite{DUNCKER2021163}. This procedure will be valid for \textbf{regimes} of low neuronal activity. We are not saying that the linear approximation is valid \textbf{only} in this regime, Only stating that in this regime, the intensity of neuronal activity is so weak that there is a formal procedure that justifies the linear approximation. In fact, this approximation was also used in the case of long times. We understand that the reason for such a thing can be justified from the differential equation and affirm that it is correct to assume a certain \textit{neuronal not saturation} in the long term. By long-term, we mean that over time, both because the matrix $\pmb{A}$ (which is diagonalizable) is such that all its eigenvalues have a real part less than 0 (this is what is called asymptotic stability) or it can always be considered that the neuronal activation function, which takes the weighted sum of the activity signals of each neuron, prepares the whole situation. In any case, the final destination of the neuronal activity is terminal, this is further ensured by the second principle of thermodynamics, in the sense that every entity at some point will have very poor neuronal activity. In this way, under linearization we will have $\pmb{F}\pmb{(}\pmb{h}(t),\pmb{x}(t)\pmb{)}\simeq -\pmb{\lambda h}(t) + \pmb{w\,h}(t)+\widetilde{\pmb{w}}\,\pmb{x}(t) $ thus the differential equation \eqref{din redd} takes the form
\begin{equation}\label{A B}
\pmb{\dot{h}}(t)=\pmb{A}\,\pmb{h}(t)+\widetilde{\pmb{w}}\,\pmb{x}(t),
\end{equation} 
where $\pmb{A}=\pmb{w} -\pmb{\lambda}$. We are interested in studying the activity of each neuron $\{h_k\}_{k\in I_n}$ subject to two types of interactions due to the interconnection: \textit{inter-neuronal} connection given the weights matrix $\pmb{w}$ and a series of external excitation $\pmb{x}$ affecting each neuron through $\widetilde{\pmb{w}}$ matrix. Linear systems are made particularly attractive by the
fact that their asymptotic behaviour can be understood
in terms of the eigenvalues and eigenvectors of $\pmb{A}$ \cite{DUNCKER2021163}. 

\section{Electrical Networks}

We will identify the system described by \eqref{din red}, in particular its linearized form \eqref{A B} with an electrical network. An electrical network (EN) is understood as the composition of an oriented graph, where each of its arcs has two associated functions of time: current and voltage, these functions are linked by Kirchhoff's laws and by the arc relations that arise from the graph that represents it and the interconnected electrical elements, e.g. resistors, inductors, capacitors, etc. \cite{Bala}. 
Whenever the method of nodes, loops or pairs of nodes \cite{Bala} is used to find the dynamics of the network, systems of linear differential equations of first order or integro-differential equations will be obtained. To be able to use these methods, it is necessary to know the electrical network entirety, i.e. the elements that compose it and their arrangement and interconnection. Since we do not have this information, the solution must be structured on a general circuit, i.e. without taking into account its graph, or the elements in each of its arcs. Therefore it is not possible to apply any of the network analysis methods until we find a particular network and its arc elements. The way to solve this apparent circular problem requires not the \textit{analysis} of one circuit, but the \textit{synthesis} of all circuits in a given preferential family. Such requirements can be fixed from other assumptions which we will see below. 

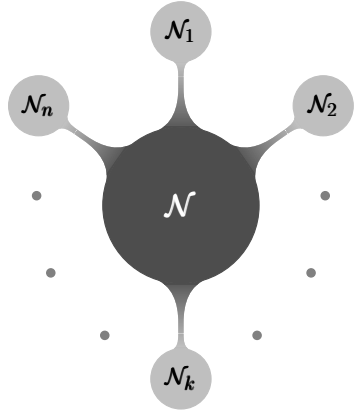
\begin{figure}
\centering
\resizebox{0.38\textwidth}{!}{%
\begin{tikzpicture}
  \path[mindmap,concept color=black!70,scale=1.3]
    node[concept,minimum size=4.5cm, color=black!70 ,text=white,scale=1.3] 
    {\Huge $\pmb{\mathcal{N}}$}
    [clockwise from=90]
    child[concept color=gray!50!, text=black] 
    {node[concept] {\Huge $\pmb{\mathcal{N}_1}$}}  
    child[concept color=gray!50!,text=black,grow=35] 
    {node[concept] {\Huge $\pmb{\mathcal{N}_2}$}}
    child[concept color=gray!50,text=black,grow=-90] { node[concept]
    {\Huge $\pmb{\mathcal{N}_k}$}}
    child[concept color=gray!50!, text=black,grow=145] 
    {node[concept] {\Huge $\pmb{\mathcal{N}_n}$} };
\draw[gray,ultra thick,circle,fill=gray] (5.4cm,0.36cm) circle(0.15cm);
\draw[gray,ultra thick,fill=gray] (4.87cm,-2.53cm) circle(0.15cm);
\draw[gray,ultra thick,fill=gray] (2.85cm,-4.87cm)   circle(0.15cm);
\draw[gray,ultra thick,circle,fill=gray](-5.4cm,0.36cm) circle(0.15cm);
\draw[gray,ultra thick,fill=gray] (-4.87cm,-2.53cm) circle(0.15cm);
\draw[gray,ultra thick,fill=gray] (-2.85cm,-4.87cm)   circle(0.15cm);
\end{tikzpicture}
}
\caption{\footnotesize{There are $n$ dipole networks, denoted by $\{\mathcal{N}_k\}_{k\in I_n}$ interconnected through an \textit{interaction} network $\mathcal{N}$.}}\label{nTopology}
\end{figure}

We intend to distinguish or identify \textit{generalized coordinates} in the electrical network such that the equations \eqref{A B} are satisfied. Since that equation has $n$ degrees of freedom,  therefore we distinguish $n$ local regions in the network from which voltages or currents can be measured. These regions are called ports: a pair of terminals that allow to exchange energy with the surroundings and have a given port-voltage and port-current. We conclude such an electrical network has $n$ independent ports perfectly identified. The general structure that we propose is an electrical network composed of $n$ dipoles listed as $\{\mathcal{N}_k\}_{k\in I_n}$, also called one$-$port networks, interconnected through a $n-$port interaction network  $\mathcal{N}$ \cite{Bala,Carlin}. We see that each of one$-$port network introduces a port$-$voltage or port$-$current, that will be corresponds to a coordinate, $h_k$ from $\pmb{h}$ in \eqref{A B}, see figure \ref{nTopology}. Since $\pmb{w}$ is time$-$independent, then the dynamics of the power grid must be invariant under time translations, this implies that this part of the $\mathcal{N}$ network does not contain internal generators. In fact, by observing $\eqref{A B}$ we can conclude that such generators are well identified with the excitation signals $\pmb{x}$.  The initial conditions will be given on each dipole (or $one-$port network), and also in the external excitation $\pmb{x}$, the energy initially contained by the interaction network $\mathcal{N}$ can be nonzero. In this way, the theory of multi$-$port networks can be perfectly used to synthesise $\mathcal{N}$ as an \textit{active electrical network} \cite{Bala}. In this scenario is possible to define a transfer matrix function of $\mathcal{N}$ written as the quotient between the Laplace transformation to certain output signals and the Laplace transformation of certain input signals.  Depending on which signals are considered as inputs and outputs, there are four general representations:  transmission, impedance, admittance, or hybrid.  Note that the last case of hybrid representations of $\mathcal{N}$ is ruled out,  otherwise, the input and output signals, voltages and currents, from different ports would be mixed, thus losing the possibility that each network $\{\mathcal{N}_k\}_{k\in I_n}$ represents, \textit{per se}, one and only one of the coordinates $\{h_k\}_{k\in I_n}$ described by  \eqref{A B}. In conclusion, we have interested in transmission, impedance or admittance representation of the $n-$port network $\mathcal{N}$. The energy is initially provided by the list of one$-$port networks $\{\mathcal{N}_k\}_{k\in I_n}$. The corresponding dynamics of an electric network is defined by the appropriate use of the Kirchhoff rules that take care of the topology of the network, schematically represented in figure \ref{nTopology}.
We have said that the generalized coordinates will be port-voltages or port-current of each of the $n$ networks of the list 
$\{\mathcal{N}_k\}_{k\in I_n}$. Therefore, the $n-$port network $\mathcal N$ acts as an interaction in the sense that it physically interconnects the $n$ dipoles networks. So the non-interaction case corresponds to disconnecting the $\mathcal{N}$ network. In a RNN this interaction-free responds to the fact that the weights matrix satisfies  $\pmb{w}=\pmb{0}$ and the external excitation $\pmb{x}=\pmb{0}$ or its weithgts matrix $\widetilde{\pmb{w}}=\pmb{0}$, so that following \eqref{A B} each neuron has an activity signal given by $h_k(t){=} \alpha_k e^{-\lambda_k\, t}$, for a given $\lambda_k {\in} \mathbb{R}$. In other words neurons do not \textit{see} each other. Mathematically, this is due because the evolution equation \eqref{A B} take the form
\begin{equation}\label{non-int net}
\pmb{\dot{h}}(t)=-\pmb{\lambda h}(t),
\end{equation} 
and given the matrix $\pmb{\lambda}$ is diagonal the system of equations is uncoupled, the activity of the neurons is relegated to its initial condition and to a behavior that decays exponentially with a characteristic time matrix.  

\begin{figure}
\begin{center}
\resizebox{0.63\textwidth}{!}{%
\begin{circuitikz}[scale=.9,american currents, american voltages, color=black,very thick]
\tikzset{font=\fontsize{14}{0}\selectfont}
\ctikzset{color=red!60!black,bipoles/thickness=1,resistors/scale=0.8,capacitors/scale=0.8}
 \draw[color=red!60!black](0,4) to[vsource,invert, l_=$v_k(0)$,red!60!black,color=red!60!black] (0,7)
 to[short, -*] (2,7)
 to[R=$R_k$,color=red!60!black] (2,4) -- (0,4);
 \draw[color=red!60!black] (2,7) -- (4,7)  to[C=$C_k$]
 (4,4) to[short, -*] (2,4);
\draw[color=red!60!black] (2,7) 
 to[short, *-*] (4,7) to [short, -*](4,7)node[squarepole,scale=1.5,label={right:$a_k$}] {};
 \draw[color=red!60!black] (2,4) 
 to[short, *-*] (4,4) to [short, -*](4,4)node[squarepole,scale=1.5,label={right:$b_k$}] {};
  \end{circuitikz}
  \qquad
\begin{circuitikz}[scale=.9,american currents, american voltages, color=black,very thick]
 \ctikzset{bipoles/thickness=1,color=blue!40!black,resistors/scale=0.8}
 \tikzset{font=\fontsize{14}{0}\selectfont}
  \draw[color=blue!40!black](0,0) 
  to[isource,l_=$i_k(0)$,blue!40!black] (0,3) 
  to[R,l_=$R_k$] (4,3); %
  \draw[color=blue!40!black](0,0) 
  to[L=$L_k$] (4,0); 
   \draw[color=blue!40!black] (4,3) 
  to[short, -*] (4,3) node[squarepole,scale=1.5,label={right:$a_k$}] {};
   \draw[color=blue!40!black] (4,0) 
  to[short, -*] (4,0) node[squarepole,scale=1.5,label={right:$b_k$}] {};
  \draw[color=blue!40!black] (4,0) 
  to[short, -*] (4,3)  {};
 \end{circuitikz}
}
\end{center}
 \caption{\footnotesize{Alternatives  networks for each $\mathcal{N}_k$ in non-interaction case, i.e.  a terminal dipole network of the list $\{\mathcal{N}_k\}_{k\in I_n}$ and its initial excitation, depending on which signal: voltages $v_k$ (left and \textcolor{red!60!black}{red})  or currents $i_k$ (right and \textcolor{blue!40!black}{blue}) is chosen  to describe the coordinate $h_k$ of \eqref{A B}. }}\label{dipoloz}
\end{figure}
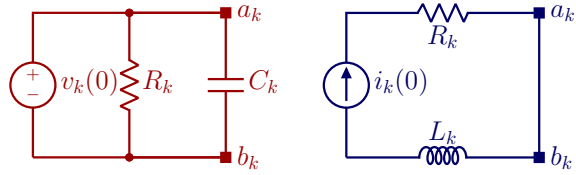

In order to compare directly with the result from the network synthesis method, let's apply the Laplace transform ($\mathscr{L}$) of the above linear differential equation  \eqref{non-int net} and taken the $k-$component of $\pmb{h}(t)$ denoted by $h_k(t)$, then $H_k(s){=}\,h_k(0)/(s+\lambda_k)$, where     $H_k(s){=}\mathscr{L}\pmb{(}h_k(t)\pmb{)}_{(s)}$ and a nonnegative matrix $\pmb{\lambda}=diag(\cdots, \lambda_k,\cdots)$. Note that each $H_k(s)$ is conceived as a characteristic function of a one-port $R_k{\parallel} C_k$ in parallel  or one-port $R_k{\boxplus} L_k$ in series \cite{Bala}, these circuits are to be dual to each other. The identification ($\widehat{=}  $) with the equation \eqref{A B} is as follows: $\lambda_k^{-1}\widehat{=} R_kC_k$ and $h_k\widehat{=} v_k$ or $\lambda_k^{-1}\widehat{=} R_k/L_k$ and $h_k\widehat{=} i_k$. It should be noted that in each case the variable chosen is common to all of its elements: voltages and current for the parallel and series cases, respectively. Figure \ref{dipoloz} summarize this situation.
Note that the voltage source $v_k(0)$ and the current source $i_k(0)$ in figure \ref{dipoloz} represent not only the initial condition but also mention that the initial energy is stored in the reactive elements. From an electromagnetic point of view, the initial potential difference in the capacitor $C_k$ refers to the stored electrical energy given by $\frac{1}{2} C_k v_k^2(0)$. While the initial current in the inductor $L_k$ refers to the stored magnetic energy given by $\frac{1}{2} L_k i_k^2(0)$. That is, both reactive elements are the initial source and thus provide the initial condition in each case.

As we have said, the interaction of the components of neuronal activity vector $\pmb{h}$ is due to an \textit{internal} connection between neurons regulated by the internal weights matrix $\pmb{w}$ and to an \textit{external} connection regulated by the excitation $\pmb{x}$ and the external weights matrix $\widetilde{\pmb{w}}$. 
To establish the RNN correspondence with the electrical networks in this linear context, we need to connect the interaction network $\mathcal{N}$ in order to identify it with the matrices $\pmb{w}$ and  $\widetilde{\pmb{w}}$ and the external excitation $\pmb{x}$.

We claim that for a given a recurrent neural network regulated by \eqref{A B}, there are two electrical circuits, $\parallel$-\textit{parallel}  and $\boxplus$-\textit{series} networks, that are dual to each other and reproduce the dynamics proposed by \eqref{non-int net}. 

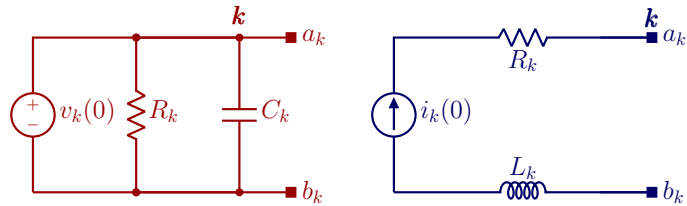
\begin{figure}[ht!]
\begin{center}
\resizebox{0.75\textwidth}{!}{%
\begin{circuitikz}[scale=1,american currents, american voltages, color=black,very thick]
\tikzset{font=\fontsize{14}{0}\selectfont}
\ctikzset{color=red!60!black,bipoles/thickness=1,resistors/scale=0.8, capacitors/scale=0.8}
 \draw[color=red!60!black](0,4) to[vsource,invert, l_=$v_k(0)$,red!60!black,color=red!60!black] (0,7)
 to[short, -*] (2,7)
 to[R=$R_k$,color=red!60!black] (2,4) -- (0,4);
 
 \draw[color=red!60!black] (2,7) -- (4,7) node[label={above:$\pmb{k}$}] {} to[C=$C_k$] 
 (4,4) to[short, -*] (2,4);
\draw[color=red!60!black] (2,7) 
 to[short, *-*] (4,7) to [short, -*](5,7)node[squarepole,scale=1.5,label={right:$a_k$}] {};
 \draw[color=red!60!black] (2,4) 
 to[short, *-*] (4,4) to [short, -*](5,4)node[squarepole,scale=1.5,label={right:$b_k$}] {};
  \end{circuitikz}
    \qquad
\begin{circuitikz}[scale=1,american currents,american voltages, color=black,very thick]
 \ctikzset{bipoles/thickness=1,color=blue!40!black,resistors/scale=0.8}
 \tikzset{font=\fontsize{14}{0}\selectfont}
  \draw[color=blue!40!black](0,0) 
  to[isource,l_=$i_k(0)$,blue!40!black] (0,3) 
   to[R,l_=$R_k$] (5,3); %
  \draw[color=blue!40!black](0,0) 
  to[L=$L_k$] (5,0); 
   \draw[color=blue!40!black] (4,3) 
  to[short, -*] (5,3) node[squarepole,scale=1.5,label={right:$a_k$},label={above:$\pmb{k}$}] {};
   \draw[color=blue!40!black] (4,0) 
  to[short, -*] (5,0) node[squarepole,scale=1.5,label={right:$b_k$}] {};
 \end{circuitikz}
}
\end{center}
 \caption{\footnotesize{Alternatives networks for $\mathcal{N}_k$, i.e.  a terminal dipole network of the list $\{\mathcal{N}_k\}_{k\in I_n}$ and its initial excitation, depending on which signal: now are port$-$voltages $v_k$ (left and \textcolor{red!60!black}{red}) or port$-$currents $i_k$ (right and \textcolor{blue!40!black}{blue}) is chosen  as a coordinate $h_k$ of \eqref{A B}.  }}\label{dipolos}
\end{figure}

Note that the radical difference of each subnetwork $\mathcal{N}_k$ between the figure \ref{dipoloz} and \ref{dipolos} is that the common signal $v_k$ ($i_k$) for the parallel (serial) case is transmitted to the network $\mathcal{N}$; so each pair of terminals $(a_k,b_k)$ are arranged to ensure this effect. To complete the electrical configuration of the complete network in \ref{nTopology}, each pair of terminals $(a_k,b_k)$ conforms a port that is connected to the $k-$port of $\mathcal{N}$. Depending on whether you choose to use a description in terms of voltages or currents, you will have to use an admittance or impedance representation for the associated $\mathcal{N}$ network.

To fix ideas we choose the  \textit{parallel network} and port$-$voltages of each subnetwork of the list $\{\mathcal{N}_k\}_{k\in I_n}$ as generalized coordinates, $\pmb{v}=(v_1,{\cdots},v_k,{\cdots},v_n)$, at left on the figure \ref{dipolos}. For each $k{\in}I_n$, a subnetwork $\mathcal{N}_k$ is a $R_k\Vert C_k$ tandem circuit, which is connected to the $k-$port of the network $\mathcal{N}$ as showed in figure \ref{nTopology}. The $k-$node conform the $k-$port given by the pair of terminals $(a_k,b_k)$, applying  Kirchhoff's first law at this $k-$node:  $i_{\scriptscriptstyle{C_k}}+i_{\scriptscriptstyle{R_k}}-i_k{=}0$, using that $i_{\scriptscriptstyle{R_k}}{=}v_k/R_k$ and $i_{\scriptscriptstyle{C_k}}{=}C_k d_tv_k$, thus $C_k\dot{v}_k(t)+R_k^{-1}v_k(t)- i_k(t)=0$, performing a Laplace transform ($\mathscr{L}$) then $
C_ksV_k(s)-C_kv_k(0)+R_k^{-1} V_k(s)-I_k(s)=0$. The last equation can be expressed in matrix form as 
\begin{equation}\label{aha}
s\pmb{V}(s)-\pmb{v}(0)+\pmb{\mathrm{\Lambda}}\:\pmb{V}(s)-\pmb{C}^{-1}\pmb{I}(s)=\pmb{0},
\end{equation}
where $\pmb{C}=diag(C_1,\cdots,C_n)$ and $\pmb{R}=diag(R_1,\cdots,R_n)$, the matrix $\pmb{\mathrm{\Lambda}}{=}(\pmb{RC})^{-1}$ contains the inverse of the characteristics times of each $R_k \Vert C_k$ subnetworks. In such a way, it all comes down to synthesizing the $\mathcal{N}$ network in the sense of the \cite{Bala} circuit theory, in order to obtain a relation between the port-currents $i_k$ and the port-voltages $v_k$. If the RNN does not have external excitation then the synthesis of the EN, $\mathcal{N}$, allows to express $\pmb{I}(s)=\pmb{Y}(s)\pmb{V}(s)$, we have used the admittance representation of $\mathcal{N}$. Applying the inverse Laplace ($\mathscr{L}^{-1}$) transform to obtain the equation in the time domain
\begin{equation}\label{aha2}
\pmb{\dot{v}}(t)+\pmb{\mathrm{\Lambda} v}(t)-\pmb{C}^{-1}\mathscr{L}^{-1}[\pmb{Y}(s)\pmb{V}(s)]{(t)}{=}\pmb{0}.
\end{equation}
The matrix elements of $\pmb{Y}(s)$ are rational functions: quotients of polynomials in $s$. A necessary and sufficient condition for that the equation \eqref{aha2} has the  form of \eqref{A B} is $\pmb{Y}(s)=\pmb{\alpha}$, where the constant matrix $\pmb{\alpha}$ of \textit{conductances} can be synthesized using the general method exposed in \cite{Carlin}. There is non restrictions about the symmetry of the matrix $\pmb{\alpha}$, in other words if we are interested in considering $\pmb{\alpha}$ that is not necessarily symmetric, then the network $\mathcal{N}$ is said to be non-reciprocal. This implies that it can be synthesized using \textit{gyrators} \cite{Carlin}. Comparing equations \eqref{aha2} and \eqref{A B}  we obtain the identification $\pmb{w}\hat{=}\pmb{C}^{-1}\pmb{\alpha}{=:}\pmb{\mathrm{\Omega}}$.

If we consider the complete description of RNN with the external excitation $\pmb{x}$ then the electrical network  $\mathcal{N}$ must be synthesized by 
$\pmb{I}(s)=\pmb{\alpha}\pmb{V}(s)+\pmb{\beta}\pmb{U}(s)$, where $\pmb{U}(s){=}\mathscr{L}[\pmb{u}(t)](s)$ and $\pmb{u}(t)=(u_1,\cdots,u_m)$ are the voltages sources that act as the external excitation $\pmb{x}$, in these case the weight matrix $\widetilde{\pmb{w}}\hat{=}\pmb{C}^{-1}\pmb{\beta}{=:}\widetilde{\pmb{\mathrm{\Omega}}}$. 

A similar procedure can be repeated in the impedance representation of the network $\mathcal{N}$ simply by interchanging the following quantities: voltages by currents, inductances by capacitances, conductances by resistances in order to obtain identical equations to \eqref{A B} so that now the generalized coordinates are the port$-$currents $\pmb{i}$.

The procedure we have described can be summarized in the following steps:
\begin{enumerate}
\item Propose the general topology of $n$ sub-networks $\{\mathcal{N}_k\}_{k \in I_n}$ connected to a $\mathcal{N}$ network, hoping to be able to associate each artificial neuron in the RNN with a sub-network of $\{\mathcal{N}\}_{k \in I_n}$.
\item Identify the case of no interaction in both systems, in this case, RNN and EN.
\item Look for a dynamic quantity common to the, and representative of the sublattice $\mathcal{N}_k$ that corresponds to $h_k$. Note that it needs to be common and representative to capture that shared in $\mathcal{N}_k$ and to be able to identify the dynamics of each $\mathcal{N}_k$ with that of each $h_k$.
\item In the case considering interaction, this common and representative information of each $\mathcal{N}_k$ must be transferred to $\mathcal{N}$. This is achieved by transferring the potential difference ($v_k$) or the current flowing through the arc $a_k$ and $b_k$ ($v_k$) or the current between $a_k$ and $b_k$ (by previously opening such terminals) to the $k-$gate of $\mathcal{N}$.

\end{enumerate}

The dynamics of the Electrical Network follows the differential equation
\begin{equation}\label{sim A B}
\pmb{\dot{v}}(t)=-\pmb{\mathrm{\Lambda} v}(t)+\pmb{\mathrm{\Omega}}\,\pmb{v}(t)+\widetilde{\pmb{\mathrm{\Omega}}}\,\pmb{u}(t) {=}\pmb{0},
\end{equation}
we summarize the identification of  the elements of an RRN \eqref{A B} and the class of Electrical Networks \eqref{sim A B} under study: $\pmb{h}{\hat{=}}\pmb{v}$, $\pmb{\lambda}{\hat{=}}\pmb{\mathrm{\Lambda}}$,
$\pmb{\omega}{\hat{=}}\pmb{\mathrm{\Omega}}$,  
$\widetilde{\pmb{\omega}}{\hat{=}}\widetilde{\pmb{\mathrm{\Omega}}}$ and $\pmb{x}{\hat{=}}\pmb{u}$.

For the nonlinear case, where the activation function $\sigma$ plays am essential role, then we must consider the use of nonlinear amplifiers  synthesis method with feedback in $\mathcal{N}$.

\section{Discussion}

It is well known that any finite-time trajectory of a given $n-$dimensional dynamical system can be approximately realized by the internal state of the output units of a continuous-time recurrent neural network with $n$output units. From this idea, and with the advance of the last ten years which includes current computing algorithms, the new hardware and the theory of neural networks, we have enormous developments in various areas related to natural language, dynamical systems, neuroscience and time series analysis.

While it may seem like an unnecessary step, being able to formalize and ground fundamental connections that are directly used from the very beginning allows us to learn more about systems in the process. It allows us to contextualize the types of circuits used and identify their characteristics and the parameters of the recurrent networks.

We have also carried out a review that allows us to present the current state of the art in the field of recurrent neural networks. We can perform simulations where we have an accurate representation of the phenomena associated with these systems. We have now the confidence that the representation is mathematically equivalent to the equations that represent the network.

\section{Conclusions and Future Work}

Since the idea of the current work was to formalize this equivalence implemented in the last 30 years, the objective was the development of such formalism in present paper. We have presented a procedure  summarized in 4 steps to identify the elements of the electrical network with elements of the equation that represents the recurrent neural network.
We think that a future work could address how to include specific conditions that neural networks must meet, such as Dale's law, or other constraints of biological origin and how they can affect the parameters of the circuits that emulate the networks.

\subsubsection{Acknowledgements}
Present work was supported by FIDESOL, CONICET and UNQ.

%
%
%
\bibliographystyle{splncs04}

\bibliography{mybibfile.bib}

\end{document}